\documentclass[10pt, a4paper]{article}

\usepackage{CVPR2024-Low-light-Object-Detection} 
\usepackage{indentfirst}
 \usepackage{hyperref}
\title{Low-light Object Detection}
\name{Pengpeng Li\textsuperscript{1} , Haowei Gu\textsuperscript{1} , Yang Yang\textsuperscript{1}* \\}

\address{Nanjing University of Science and Technology \\ \\
\{lirabbit2s@gmail.com, 418959883@qq.com, yyang@njust.edu.cn\}}

\abstract{
In this competition\footnote {Corresponding author:Yang Yang(yyang@njust.edu.cn)} we employed a model fusion approach to achieve object detection results close to those of real images. Our method is based on the CO-DETR model, which was trained on two sets of data: one containing images under dark conditions and another containing images enhanced with low-light conditions. We used various enhancement techniques on the test data to generate multiple sets of prediction results. Finally, we applied a clustering aggregation method guided by IoU thresholds to select the optimal results.
 \\ \newline  }

\begin{document}

\maketitleabstract

\section{Competition introduction}
\par
With the development of machine learning, various deep learning models continue to emerge.An open-world SSL method for self-learning multiple unknown classes, surpassing state-of-the-art methods on various benchmarks, including ImageNet-100.\cite{xi2023robust}  the authors apply Semantic Sharing to ensure consistent embedding for cross-modal retrieval by training each modality's classification performance on a shared self-attention based model\cite{yang2021rethinking} the authors propose Transductive Federated Learning (TFL) to address the challenge of making inferences for newly-collected data in a privacy-protected pilot project.\cite{li2023mrtf}.This competition belongs to CV field.

The background of this competition revolves around the detection of objects in images captured under extremely low-light conditions.\cite{zou2023object} The dataset consists of eight types of objects, ranging from bicycles and bottles to tables and other everyday items. In these low-light scenarios, objects may often appear stacked or partially obscured, adding complexity to the detection task. \cite{hong2021crafting}This stacking phenomenon presents a significant challenge as it requires the detection algorithm to accurately identify and delineate overlapping objects, contributing to the overall difficulty of the task.
\begin{figure}[!ht]
\begin{center}
\includegraphics[scale=0.2]{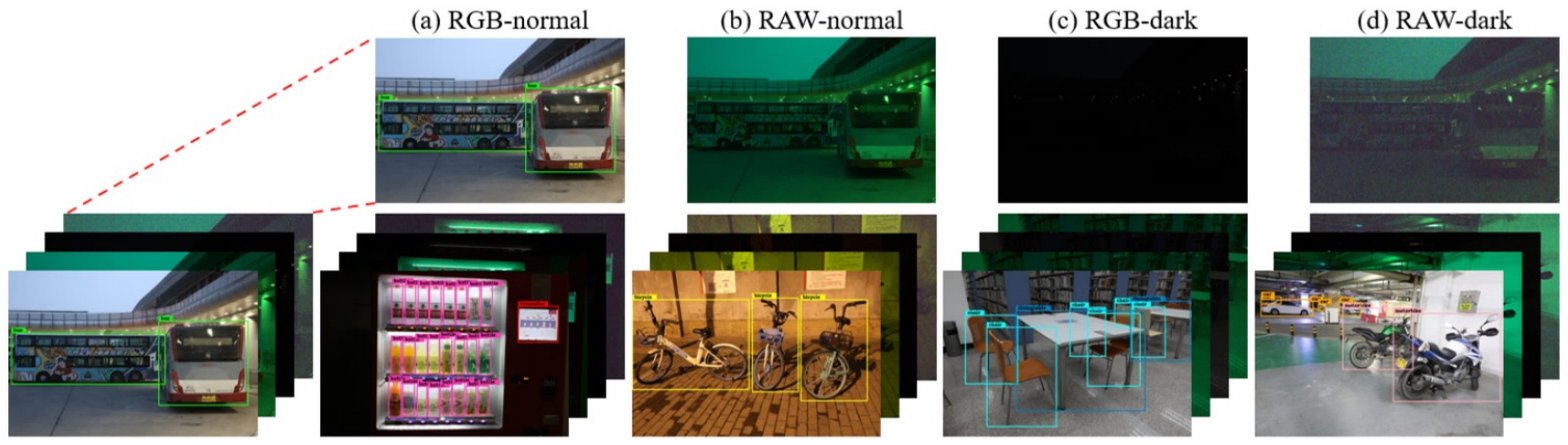} 
\caption{The example of competition dataset}
\label{fig.1}
\end{center}
\end{figure}

\section{Model structure}

The transformers structure is used in many places\cite{xi2023robust} introduce Context-Aware Transformer (CAT) with a self-supervised learning framework to effectively segment videos into cohesive scenes by comprehensively considering the complex temporal structure and semantic information, achieving state-of-the-art performance on the MovieNet dataset with a 2.15 improvement on AP for scene segmentation.\cite{meng2024multiscale}propose Multiscale Grouping Transformer with Contrastive Language-Image Pre-training (CLIP) latents (MG-Transformer).

The foundation of our model is built upon the CO-DETR architecture. DETR, standing for "Detection Transformer," is a transformer-based model in the field of object detection, known for its end-to-end characteristics.\cite{carion2020end} CO-DETR extends DETR by incorporating a conventional object detector as the detection head. \cite{zong2023detrs}It also improves upon DETR's loss function by increasing the number of positive samples, which enhances model convergence.

The CO-DETR architecture comprises an encoder-decoder structure. The encoder utilizes a transformer-based architecture to process input images and extract features. Meanwhile, the decoder generates object queries and refines object predictions. This combination enables CO-DETR to capture both global and local context information, essential for accurate object detection.

Furthermore, CO-DETR introduces several enhancements over traditional DETR:

\begin{figure}[!ht]
\begin{center}
\includegraphics[scale=0.2]{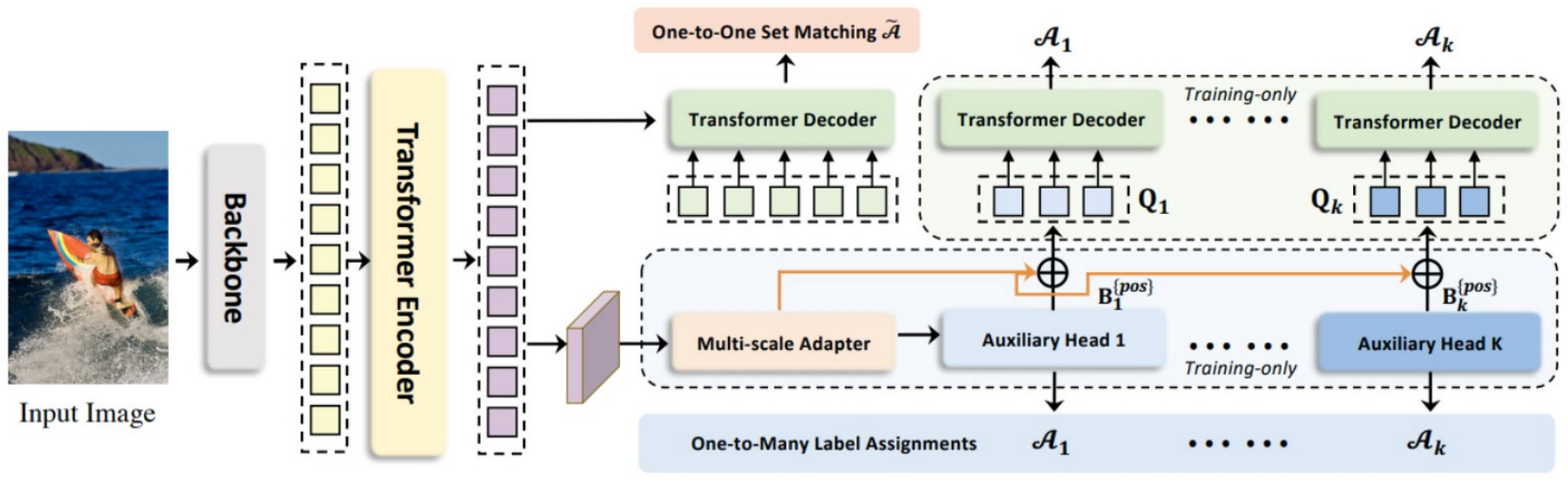} 
\caption{The structure of CO-DETR}
\label{fig.2}
\end{center}
\end{figure}

\begin{enumerate}
\item \texttt{Integration of Conventional Detector: CO-DETR incorporates a conventional object detector into the architecture, enhancing its ability to handle complex detection tasks.}
\item \texttt{Improved Loss Function: The loss function of CO-DETR is refined to improve model convergence and performance.}
\item \texttt{Increased Positive Samples: By augmenting the number of positive samples, CO-DETR enhances its ability to detect objects effectively, particularly in challenging scenarios.}

In summary, our model architecture, based on CO-DETR, combines the strengths of transformer-based models with conventional object detection techniques to achieve robust and accurate object detection performance in extremely low-light environments. By leveraging the transformer architecture, our model captures both global and local context information, essential for detecting objects in challenging lighting conditions. Additionally, the incorporation of traditional object detection techniques in the CO-DETR framework enhances its adaptability to low-light scenarios. Through this comprehensive approach, our model demonstrates superior performance in accurately detecting objects, even in the most challenging lighting conditions.

\end{enumerate}

\section{Training strategy}

In this competition, our objective is to detect objects in extremely low-light environments.
To address this challenge, we employ a comprehensive training strategy that leverages model fusion and specialized techniques. We train three separate object detection models using dark images, images enhanced for low-light conditions using the IAT model, and images augmented with the NUScene dataset. During testing, we apply various transformations to the test images and use a clustering approach to fuse the predictions. Through this strategy, we aim to achieve robust and accurate object detection results, capable of handling diverse lighting and scene conditions.

\subsection{IAT}
In addition to CO-DETR, we incorporate the Instance-Adaptive Transformer (IAT) model into our architecture. \cite{cui2022you} introduces instance-level adaptability, enhancing the model's ability to handle in low-light environments.
In low-light conditions, details of objects may be obscured or less distinguishable. The IAT model dynamically adjusts attention weights based on the specific characteristics of each object instance. This allows the model to focus more on relevant features even in challenging lighting conditions.

\subsection{Different models}

Multiple models working together to process data is common in machine learning  \cite{xiang2017modal} propose a Pre-trained Multi-Model Reuse approach (PM2R) utilizing potential consistency spread on different modalities, enabling efficient combination of pre-trained multi-models without re-training and addressing the main issue of final prediction acquisition from the responses of multiple pre-trained models in the Learnware framework. \cite{yang2019comprehensive} propose a Comprehensive Multi-Modal Learning (CMML) framework to address the challenge of divergent modalities in real-world multi-modal data, striking a balance between consistency and diversity using instance-level attention and novel regularization techniques, demonstrating superior performance on real-world datasets.

The use of three different datasets allows each model to focus on different aspects of the image characteristics. Specifically, the model trained on dark images captures features relevant to low-light environments, such as dimly lit scenes or nighttime settings, where objects may appear with reduced visibility. On the other hand, the model trained on images enhanced using the IAT model adapts to improved lighting conditions. The IAT model, based on the transformer architecture, effectively enhances the brightness of images captured in dark scenes, allowing the model to better perceive objects even in challenging lighting conditions. Additionally, the model trained on augmented images gains a broader understanding of scene diversity by learning from a dataset that contains a wide variety of scenes and lighting conditions. This diverse training approach equips our models with the ability to handle a range of scenarios, from low-light environments to well-lit scenes, ensuring robust performance in object detection tasks across different lighting conditions.

\subsection{TTA}
TTA stands for Data Enhancement During Testing.\cite{shanmugam2021better}
During testing, we apply various transformations to the test images to enhance the model's ability to detect objects in different scenarios. Specifically, we increase the image size from 1200x800 to 1400x1000 pixels to provide a higher resolution input, which allows the model to capture finer details and improve detection accuracy. Additionally, we adjust the image features using HSV (Hue, Saturation, Value) to change saturation and contrast. By modifying these features, we can simulate different lighting conditions and improve the model's robustness against variations in brightness and contrast across images. This preprocessing step ensures that the model can effectively detect objects across a wide range of lighting and scene conditions, ultimately leading to more reliable detection results.

\subsection{Fuse} 

Fusion is often used in machine learning to improve model results or transform models  \cite{yang2022domfn}
propose Divergence-Oriented Multi-Modal Fusion Network (DOMFN) to address the challenge of cross-modal divergence in resume assessment, adaptingively fusing uni-modal and multi-modal predictions based on the learned divergence for improved performance, with qualitative analysis revealing its superiority and explainability over baselines on real-world datasets. 
For the fusion strategy, we adopt a clustering approach. For each image's set of predictions, we group bounding boxes with Intersection over Union (IoU) values exceeding a certain threshold into clusters. This threshold is chosen to determine the level of overlap between the bounding boxes. Additionally, we use confidence scores to filter out less reliable predictions within each cluster. Specifically, we select the bounding box with the highest confidence score as the final prediction for each cluster.

By employing this clustering approach, we can effectively consolidate multiple predictions and select the most confident ones, enhancing the overall accuracy of our detection results. This method allows us to prioritize the most reliable predictions and discard redundant or less confident ones, resulting in more accurate object detection outcomes.

\begin{figure}[!ht]
\begin{center}
\includegraphics[scale=0.2]{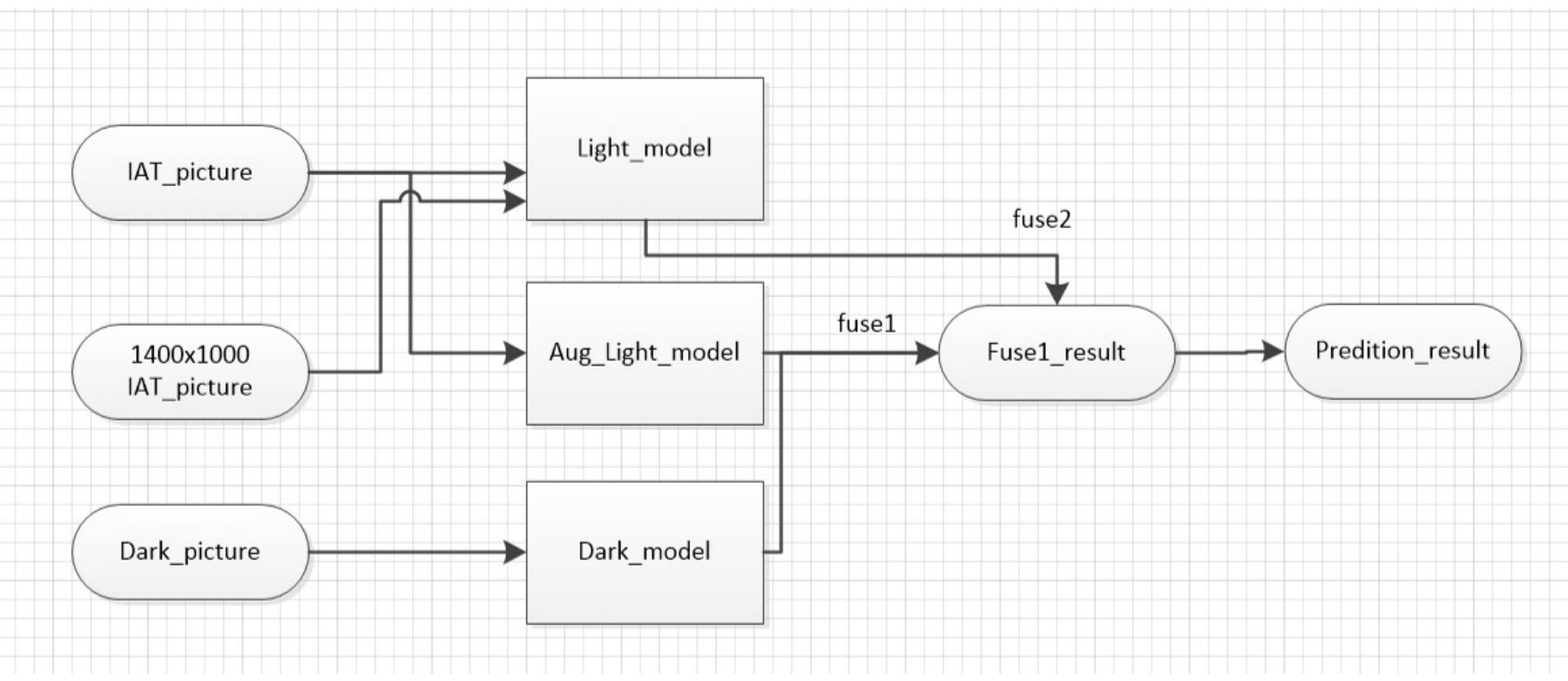} 
\caption{The process of fusion}
\label{fig.3}
\end{center}
\end{figure}

\section{Experiments}
We trained the models according to the above training steps and obtained the following results:

\begin{table}[!ht]
\begin{center}
\begin{tabularx}{\columnwidth}{|l|X|}

      \hline
      methods&result\\
      \hline
      Light-model & 0.745\\
      \hline
      Light-model+big-picture & 0.742\\
      \hline
      Light-model+aug-picture & 0.743\\
      \hline
      Dark-model & 0.732\\
      \hline
     Fusion & 0.754\\
     \hline

\end{tabularx}
\caption{The result in experiments}
 \end{center}
\end{table}

Firstly, we utilized two datasets: one containing images under dark conditions and another containing images enhanced using the IAT model. The CO-DETR model was trained separately on these two datasets to ensure adaptation to different lighting conditions.

Additionally, we employed the NUScene dataset for data augmentation to further enhance the model's generalization ability by increasing the diversity of the dataset.
During the testing phase, we applied different processing techniques to the test dataset images, including resizing and adjusting the HSV features of the images. These processing methods enabled the model to focus on the different image features under various lighting conditions and improved the model's robustness.

Finally, we employed a confidence filtering and GIoU aggregation approach to fuse the predictions of the three models.\cite{rezatofighi2019generalized} By aggregating the predictions into clusters based on GIoU values and selecting the highest-confidence prediction in each cluster, we improved the accuracy and stability of the model.
Through these experiments, we validated the effectiveness of our models in low-light environments and achieved satisfactory detection results.

\section{Conclusion}

In this study, we developed and evaluated three object detection models for detecting objects in low-light environments. By training the models on datasets containing dark images, images enhanced using the IAT model, and augmented images from the NUScene dataset, we achieved robust adaptation to diverse lighting conditions. Our experimental results demonstrate the effectiveness of our approach in improving object detection accuracy in challenging scenarios. Through careful testing and model fusion techniques, we successfully mitigated the challenges posed by low-light environments, achieving satisfactory detection results. Moving forward, our methods can be further refined and applied to real-world scenarios to enhance object detection performance in low-light conditions.

\section{Copyrights}

This document and its contents, including but not limited to text, images, and data, are the property of NJUST-KMG and are protected by copyright laws. No part of this document may be reproduced, distributed, or transmitted in any form or by any means, including photocopying, recording, or other electronic or mechanical methods, without the prior written permission of NJUST-KMG, except in the case of brief quotations embodied in critical reviews and certain other noncommercial uses permitted by copyright law.

\nocite{*}
\section{Bibliographical References}\label{sec:reference}

\bibliographystyle{CVPR2024-Low-light_Object_Detection}
\bibliography{CVPR2024-Low-light_Object_Detection}

\label{lr:ref}
\bibliographystylelanguageresource{CVPR2024-Low-light_Object_Detection}
\bibliographylanguageresource{languageresource}

\end{document}